\documentclass[conference,compsoc]{IEEEtran}
\ifCLASSOPTIONcompsoc
  \usepackage[nocompress]{cite}
\else
  \usepackage{cite}
\fi

\ifCLASSINFOpdf
  \usepackage[pdftex]{graphicx}
\else
  \usepackage[dvips]{graphicx}
\fi

\usepackage{booktabs}
\usepackage{array}
\usepackage{microtype}
\usepackage{tikz}
\usepackage{xcolor}
\usetikzlibrary{shapes,arrows.meta,positioning,fit,calc}
\usepackage[hidelinks]{hyperref}

\definecolor{methodblue}{RGB}{173,216,230}
\definecolor{methodorange}{RGB}{255,218,185}
\definecolor{methodgreen}{RGB}{144,238,144}

\begin{document}

\title{Poster: Exploring the Limits of Audio-Based Detection of Turkish Phone Call Scams}

\author{\IEEEauthorblockN{Arda Eren\IEEEauthorrefmark{1},
Micheal Cheung\IEEEauthorrefmark{1},
Youqian Zhang\IEEEauthorrefmark{1},
Grace Ngai\IEEEauthorrefmark{1} and
Eugene Yujun Fu\IEEEauthorrefmark{2}}
\IEEEauthorblockA{\IEEEauthorrefmark{1}The Hong Kong Polytechnic University, Hong Kong\\
Email: arda01.eren@connect.polyu.hk}
\IEEEauthorblockA{\IEEEauthorrefmark{2}The Education University of Hong Kong, Hong Kong}}

\maketitle

\begin{abstract}
Scam phone calls exploit vulnerable communities worldwide, yet research on detection has focused almost exclusively on English and other high-resource languages. In low-resource settings such as Turkish, detection is especially difficult, as annotated data is scarce and technological defenses remain limited. This research investigates how large language models (LLMs) can support scam detection in Turkish by introducing the first public multi-modal dataset of 100 aligned audio-transcript pairs of scam and benign conversations. We evaluate seven LLMs spanning three model families: Gemini~2.5 (Flash, Flash-Lite, Pro), GPT-4o, and Qwen (Max, Plus, Turbo), under three input conditions: raw audio, automatic speech-to-text transcripts, and transcripts refined by a native speaker. Our results suggest that transcript-based inputs consistently outperform direct audio processing, while human-corrected and uncorrected transcripts perform comparably. By centering a low-resource language and real world threat, this work highlights the urgent need for culturally and linguistically inclusive AI safety research and more robust multi-modal systems for fraud prevention.
\end{abstract}

\IEEEpeerreviewmaketitle

%-------------------------------------------------------
\section{Introduction}

Phone scams are one of the fastest growing forms of fraud, exploiting millions of victims worldwide each year. Their impact is not only financial, but also psychological, as scammers manipulate trust, authority, and urgency in ways that make detection difficult. While governments and industry have invested in countermeasures, most technical solutions remain limited to English or other high-resource languages, leaving many communities without effective protection.

At the same time, research on scam detection has been dominated by transcript-based natural language processing (NLP). These approaches assume that lexical information is sufficient to reveal intent, but in practice much of a scammer's strategy relies on affective and prosodic cues, such as intonation, stress, or vowel lengthening, that are weakened or lost when speech is reduced to text. Addressing this gap requires methods that integrate both audio and transcript modalities, especially in low-resource languages.

Turkish itself poses unique challenges: despite being spoken by millions, it is underrepresented in AI research. Turkish challenges ASR due to its agglutinative morphology and wide variation from formal to colloquial speech~\cite{tohma2020}.

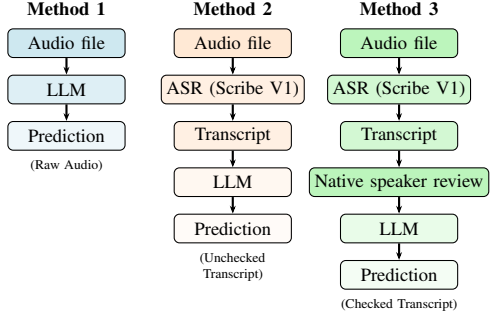
\begin{figure}[!t]
\centering
\begin{tikzpicture}[
  font=\scriptsize,
  box/.style={draw, rounded corners=2pt, text centered, minimum width=1.55cm, minimum height=0.38cm, font=\scriptsize},
  arr/.style={-{Stealth[length=3pt]}, thick},
  every node/.style={inner sep=2pt}
]

% Method 1 column
\node[box, fill=methodblue!60] (a1) at (0,0) {Audio file};
\node[box, fill=methodblue!40, below=0.22cm of a1] (m1) {LLM};
\node[box, fill=methodblue!20, below=0.22cm of m1] (p1) {Prediction};
\draw[arr] (a1)--(m1);
\draw[arr] (m1)--(p1);
\node[above=0.1cm of a1, font=\scriptsize\bfseries] {Method 1};
\node[below=0.05cm of p1, font=\tiny, text width=1.6cm, align=center] {(Raw Audio)};

% Method 2 column
\node[box, fill=methodorange!60] (a2) at (2.2,0) {Audio file};
\node[box, fill=methodorange!40, below=0.22cm of a2] (asr2) {ASR (Scribe V1)};
\node[box, fill=methodorange!40, below=0.22cm of asr2] (t2) {Transcript};
\node[box, fill=methodorange!20, below=0.22cm of t2] (m2) {LLM};
\node[box, fill=methodorange!10, below=0.22cm of m2] (p2) {Prediction};
\draw[arr] (a2)--(asr2);
\draw[arr] (asr2)--(t2);
\draw[arr] (t2)--(m2);
\draw[arr] (m2)--(p2);
\node[above=0.1cm of a2, font=\scriptsize\bfseries] {Method 2};
\node[below=0.05cm of p2, font=\tiny, text width=1.6cm, align=center] {(Unchecked Transcript)};

% Method 3 column
\node[box, fill=methodgreen!60] (a3) at (4.4,0) {Audio file};
\node[box, fill=methodgreen!40, below=0.22cm of a3] (asr3) {ASR (Scribe V1)};
\node[box, fill=methodgreen!40, below=0.22cm of asr3] (t3) {Transcript};
\node[box, fill=methodgreen!60, below=0.22cm of t3] (r3) {Native speaker review};
\node[box, fill=methodgreen!20, below=0.22cm of r3] (m3) {LLM};
\node[box, fill=methodgreen!10, below=0.22cm of m3] (p3) {Prediction};
\draw[arr] (a3)--(asr3);
\draw[arr] (asr3)--(t3);
\draw[arr] (t3)--(r3);
\draw[arr] (r3)--(m3);
\draw[arr] (m3)--(p3);
\node[above=0.1cm of a3, font=\scriptsize\bfseries] {Method 3};
\node[below=0.05cm of p3, font=\tiny, text width=1.6cm, align=center] {(Checked Transcript)};

\end{tikzpicture}
\caption{Three input conditions evaluated in this study.}
\label{fig:methods}
\end{figure}

In order to address this, we introduce the first public multi-modal dataset of Turkish scam and benign calls, consisting of 100 aligned audio and transcript pairs. We evaluate LLMs under three input conditions, as shown in Fig.~\ref{fig:methods}, to examine how transcription quality and model safety systems shape performance.

%-------------------------------------------------------
\section{Existing Solutions}

LLMs have shown promise for text-based scam detection, with studies demonstrating proficiency in identifying phishing signs in emails, though work on phone call transcripts reveals challenges such as low recall and model hallucinations.

Prior work has mostly focused on English or Chinese and often relies on edited transcripts and purely textual inputs. For instance, Shen et al.~\cite{shen2025} and Zhao et al.~\cite{zhao2018} explored mainly Chinese datasets without considering multimodal signals. These approaches overlook both the multimodal nature of scams and the realities of under-resourced languages which are critical for more inclusive AI safety research.

%-------------------------------------------------------
\section{Dataset}

The dataset contains 100 Turkish phone call recordings collected from publicly available YouTube videos, balanced between 50 scam and 50 benign calls, with all audio normalized to 16~kHz mono. 
The scam calls were sourced from YouTube videos explicitly identified as scam calls by their uploaders, and were further reviewed by a native Turkish speaker to confirm the fraudulent nature of the content.
The scam calls cover a broad range of tactics observed in Turkey, including financial and identity fraud (e.g., banking, stolen cards/IDs, government bills, top-up credit, payment issues, insurance, and prize scams) as well as service- and employment-related fraud.

%-------------------------------------------------------
\section{Proposed Method}

Our proposed approach is Method~1, in which raw audio is forwarded directly to a multimodal LLM for scam detection without any preprocessing. Methods~2 and~3 serve as comparison baselines using text-based input. In Method~2, audio is first converted to text via the Scribe~V1 ASR system and passed to the LLM without correction. In Method~3, the ASR transcript is further reviewed and corrected by a fluent Turkish speaker before being passed to the LLM.

We evaluate each of the 100 calls independently under all three input conditions, with no fine-tuning or prompt optimization performed between samples. This ensures that the results reflect standard behavior rather than a tuned pipeline.

%-------------------------------------------------------
\section{Results and Discussion}

Our analysis reveals distinct performance patterns across modalities, as shown in Table~\ref{tab:results}. Across all seven models, transcript-based inputs outperform raw audio, with mean F1 scores of 0.995 for checked transcripts and 0.992 for uncorrected transcripts compared to 0.969 for Audio. The average drop from Trans to Audio is 0.026 F1 points, while Trans and UN-Trans differ by only 0.008 on average, suggesting that ASR quality has minimal impact once a transcript exists.

A recurring failure mode in audio conditions across all model families is the refusal to process calls containing profanity or sensitive themes, such as police impersonation or extortion. Since real-world scams frequently rely on such intimidation tactics, these refusals are counted as false negatives and account for the bulk of the audio performance gap. In contrast, the same content passed as text is less likely to trigger content filters, which explains the stronger transcript results.

The audio failures point to two likely causes. First, content-filtering refusals are triggered by raw vocal signals (shouting, profanity, aggressive tone) that do not cause the same reaction in text form. Second, some models may struggle to parse overlapping speech and background noise common in real call recordings. Furthermore, during the analysis of Qwen Audio pipelines we observed hallucination artifacts likely originating from subtitled video in its training data, which could inject misleading tokens that distort downstream classification in low-resource settings.

\begin{table}[!t]
\renewcommand{\arraystretch}{1.04}
\caption{Model Performance of Different LLMs Handling the Three Types of Data}
\label{tab:results}
\centering\footnotesize
\begin{tabular}{@{}lccc@{}}
\toprule
\textbf{Models} & \textbf{Precision} & \textbf{Recall} & \textbf{F1-Score} \\
\midrule
2.5 Flash UN-Trans      & 1.00 & 0.98 & 0.99 \\
2.5 Flash Trans         & 1.00 & 1.00 & 1.00 \\
2.5 Flash Audio         & 0.94 & 0.98 & 0.96 \\
\midrule
2.5 Flash-Lite UN-Trans & 1.00 & 0.98 & 0.99 \\
2.5 Flash-Lite Trans    & 1.00 & 0.96 & 0.98 \\
2.5 Flash-Lite Audio    & 0.96 & 0.88 & 0.92 \\
\midrule
2.5 Pro UN-Trans        & 1.00 & 0.98 & 0.99 \\
2.5 Pro Trans           & 1.00 & 1.00 & 1.00 \\
2.5 Pro Audio           & 1.00 & 0.96 & 0.98 \\
\midrule
GPT-4o UN-Trans         & 1.00 & 1.00 & 1.00 \\
GPT-4o Trans            & 1.00 & 1.00 & 1.00 \\
GPT-4o Audio            & 1.00 & 1.00 & 1.00 \\
\midrule
Qwen Max UN-Trans       & 1.00 & 1.00 & 1.00 \\
Qwen Max Trans          & 1.00 & 1.00 & 1.00 \\
Qwen Max Audio          & 1.00 & 0.96 & 0.98 \\
\midrule
Qwen Plus UN-Trans      & 1.00 & 0.98 & 0.99 \\
Qwen Plus Trans         & 1.00 & 0.96 & 0.98 \\
Qwen Plus Audio         & 1.00 & 0.96 & 0.98 \\
\midrule
Qwen Turbo UN-Trans     & 1.00 & 0.96 & 0.98 \\
Qwen Turbo Trans        & 1.00 & 1.00 & 1.00 \\
Qwen Turbo Audio        & 0.98 & 0.92 & 0.95 \\
\bottomrule
\end{tabular}
\end{table}

%-------------------------------------------------------
\section{Conclusion}

This study suggests that transcript-based inputs consistently outperform raw audio for Turkish scam call detection across all evaluated LLMs, with mean F1 scores of 0.99 for both Trans and UN-Trans compared to 0.97 for Audio. Crucially,
human-corrected and uncorrected transcripts perform nearly
identically, suggesting that the cost of native speaker review
may not be justified for detection purposes.

The audio performance gap is primarily driven by content-
filter refusals on calls involving profanity, intimidation,
and police impersonation, which are the very tactics that
characterize real scams. This reveals a fundamental tension
between model safety mechanisms and practical utility in
adversarial detection tasks.

Several limitations of this study warrant consideration. The dataset, while the first of its kind for Turkish scam detection, consists of 100 calls, and the high F1 scores may partly reflect its current scope in terms of data size and diversity. Dataset artifacts such as inconsistent audio quality and uploader labeling conventions may have influenced model performance as well. Expanding the dataset and evaluating robustness across broader real-world conditions remain important directions for future work.

% Computer Society conferences typically use the plural form


\begin{thebibliography}{1}







\bibitem{shen2025}
Z.~Shen, S.~Yan, Y.~Zhang, X.~Luo, G.~Ngai, and E.~Y.~Fu, ``It warned me just at the right moment: Exploring LLM-based real-time detection of phone scams,'' in \emph{Proc. Extended Abstracts CHI Conf. Human Factors in Computing Systems}, 2025, pp.~1--7.




\bibitem{tohma2020}
K.~Tohma and Y.~Kutlu, ``Challenges encountered in Turkish natural language processing studies,'' \emph{Natural and Engineering Sciences}, vol.~5, no.~3, pp.~204--211, 2020.




\bibitem{zhao2018}
Q.~Zhao, K.~Chen, T.~Li, Y.~Yang, and X.~Wang, ``Detecting telecommunication fraud by understanding the contents of a call,'' \emph{Cybersecurity}, vol.~1, no.~1, p.~8, 2018.




\end{thebibliography}
\end{document}